\documentclass[11pt]{article}

\usepackage{placeins}
\usepackage[preprint]{acl}
\usepackage{booktabs}
\usepackage{multirow}
\usepackage{times}
\usepackage{latexsym}
\usepackage{amsmath}
\usepackage[table]{xcolor}
\usepackage{algorithm}
\usepackage{algpseudocode}
\usepackage{makecell}
\usepackage{siunitx}
\usepackage{threeparttable}
\usepackage{subcaption}
\usepackage{xcolor}
\usepackage{listings}
\usepackage{caption}
\usepackage{graphicx}
\usepackage{float}
\usepackage[table]{xcolor}
\usepackage{siunitx}
\usepackage{amssymb}
\usepackage{wasysym}

\newcommand\footnoteONLYtext[1]{
    \let \mybackup \thefootnote
    \let \thefootnote \relax
    \footnotetext{#1}
    \let \thefootnote \mybackup
    \let \mybackup \imareallyundefinedcommand}

\definecolor{chatdevbg}{RGB}{247,250,255}
\definecolor{edrbg}{RGB}{248,248,248}
\definecolor{boxframe}{RGB}{218,224,232}

\lstdefinestyle{markdownRawChatDev}{
  backgroundcolor=\color{chatdevbg},
  basicstyle=\ttfamily\tiny,
  breaklines=true,
  breakatwhitespace=false,
  columns=fullflexible,
  keepspaces=true,
  frame=single,
  rulecolor=\color{boxframe},
  framerule=0.35pt,
  xleftmargin=0.3em,
  xrightmargin=0.3em,
  aboveskip=0.5em,
  belowskip=0.5em,
  captionpos=b
}

\lstdefinestyle{markdownRawEDR}{
  backgroundcolor=\color{edrbg},
  basicstyle=\ttfamily\tiny,
  breaklines=true,
  breakatwhitespace=false,
  columns=fullflexible,
  keepspaces=true,
  frame=single,
  rulecolor=\color{boxframe},
  framerule=0.35pt,
  xleftmargin=0.3em,
  xrightmargin=0.3em,
  aboveskip=0.5em,
  belowskip=0.5em,
  captionpos=b
}

\usepackage[T1]{fontenc}

\usepackage[utf8]{inputenc}

\usepackage{microtype}

\usepackage{inconsolata}

\usepackage{graphicx}

\definecolor{DeltaRow}{RGB}{245,247,250}
 
\title{ChatDev 2.0: A No-Code Multi-Agent Platform for Developing Everything}

\author{
\textbf{Yufan Dang}{\footnotesize $^\bigstar$\textsuperscript{\textdagger}} \quad
\textbf{Shu Yao}{\footnotesize $^\clubsuit${\footnotesize $^\S$}\textsuperscript{\textdagger}} \quad
\textbf{Bowen Lai}{\footnotesize $^\spadesuit$} \quad
\textbf{Chenting Xu}{\footnotesize $^\ddagger$} \quad\\
\textbf{Ruijie Shi}{\footnotesize $^\diamondsuit$} \quad 
\textbf{Leung Wai-Shing}{\footnotesize $^\bigstar$} \quad 
\textbf{Huatao Li}{\footnotesize $^\clubsuit$} \quad 
\textbf{Chen Qian}{\footnotesize $^\clubsuit$\textsuperscript{*}} \quad 
\textbf{Zhiyuan Liu}{\footnotesize $^\bigstar$\textsuperscript{*}} \quad 
\\
{\footnotesize $^\bigstar$}Tsinghua University \quad
{\footnotesize $^\clubsuit$}Shanghai Jiao Tong University \quad 
{\footnotesize $^\S$}Shanghai Innovation Institute\quad \\
{\footnotesize $^\spadesuit$}Fudan University \quad
{\footnotesize $^\ddagger$}Peking University\quad
{\footnotesize $^\diamondsuit$}Massachusetts Institute of Technology   \\
\href{dangyf25@mails.tsinghua.edu.cn}{\texttt{dangyf25@mails.tsinghua.edu.cn}} \quad 
\href{springrainyszxr@gmail.com}{\texttt{yao.shu2004@outlook.com}}\quad \\
\href{qianc@sjtu.edu.cn}{\texttt{qianc@sjtu.edu.cn}} \quad
\href{liuzy@tsinghua.edu.cn}{\texttt{liuzy@tsinghua.edu.cn}}
\\
}

\begin{document}
\maketitle
\footnoteONLYtext{$^\dagger$Equal Contribution.}
\footnoteONLYtext{$^{\text{*}}$Corresponding Author.}

\begin{abstract}
Large language model (LLM)-based multi-agent systems (MAS) have shown strong potential for solving complex tasks, yet their development forces a tradeoff: code frameworks are expressive but engineering-intensive, while no-code builders simplify authoring but constrain agent interactions to author-defined workflows. We present \textbf{ChatDev 2.0: DevAll} (hereafter DevAll), a no-code platform for building, executing, and inspecting heterogeneous MAS that delivers both high expressiveness and ease of use. In terms of expressiveness, DevAll pairs a declarative executable graph abstraction with a cycle-aware execution engine, so that heterogeneous agents and dynamic and cyclic interactions can be represented and executed within a single framework. For ease of use, an integrated visual interface lets users author, run, monitor, and inspect MAS, including human-in-the-loop steps, entirely without writing code. Experiments demonstrate that DevAll reproduces state-of-the-art MAS across three representative tasks at competitive performance and without task-specific orchestration code, highlighting its effectiveness as a general-purpose platform for LLM-based MAS. DevAll is available at \url{https://github.com/OpenBMB/ChatDev}.
\end{abstract}
\textbf{}

\section{Introduction}

\begin{figure}[t]
\centering
\includegraphics[width=0.5\textwidth]{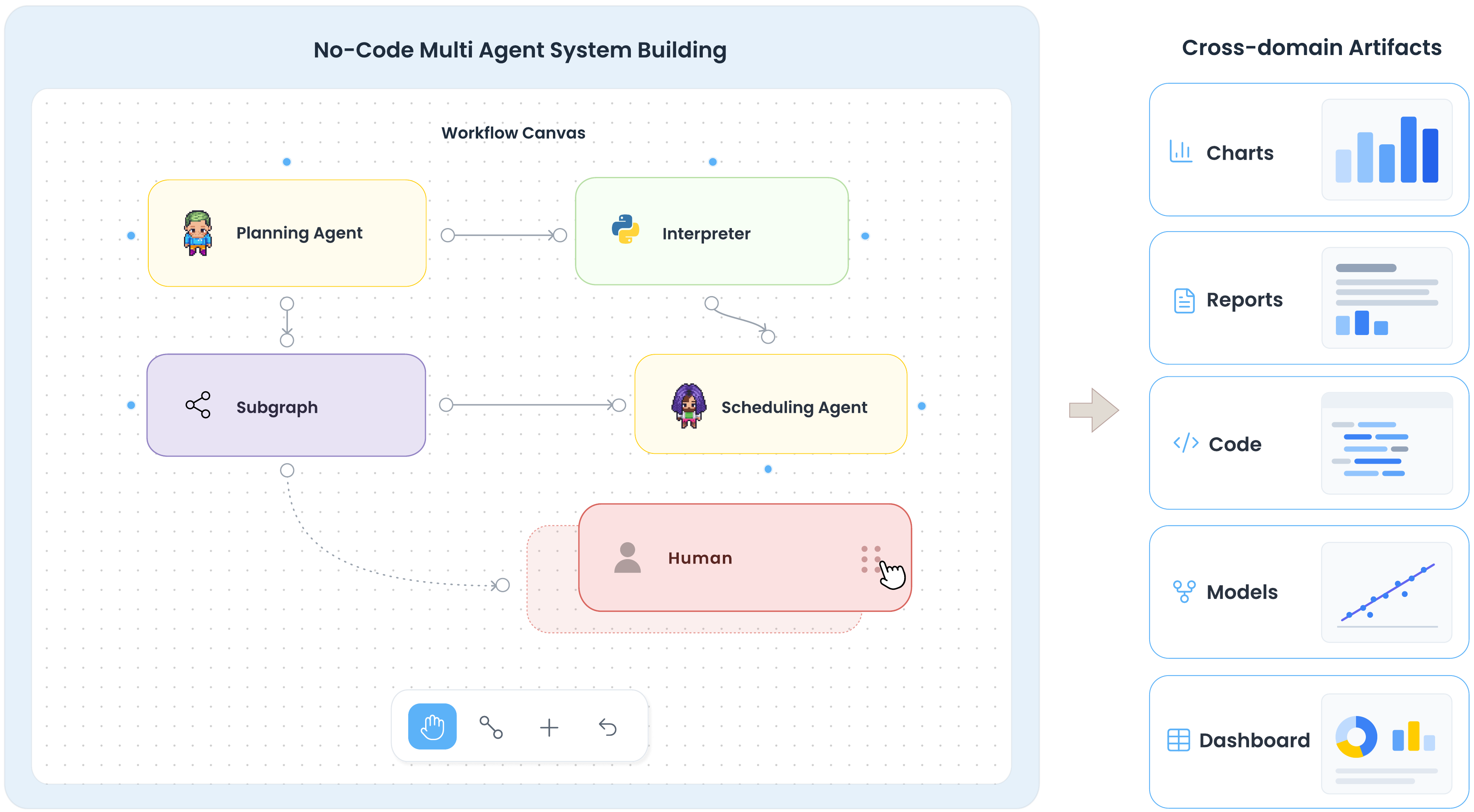}
\caption{Users drag heterogeneous components onto the canvas and connect them into executable multi-agent systems. DevAll supports cross-domain artifact generation, including charts, research reports, and so on.}
\label{fig:teaser}
\end{figure}

LLM-based multi-agent systems (MAS) coordinate specialized agents to decompose complex tasks, exchange intermediate results, and respond to runtime feedback~\citep{he2025llmbased,chen2024agentverse,wu2023autogen}. They have been applied to software engineering~\citep{qian2024chatdev,hong2024metagpt,antoniades2024swe}, deep research and information synthesis~\citep{chen2024mindsearch,huang2025manusearch}, and structured content and visualization generation~\citep{chen2026coda,yang-etal-2024-matplotagent}.

As multi-agent systems grow in scale and heterogeneity, building and maintaining them requires developers to configure agents and tools, specify communication and orchestration, manage shared state, and diagnose failures across long execution traces~\citep{dibia2024autogenstudio,agentscope,cemri2025why}. Programming-oriented frameworks such as AutoGen, MetaGPT, AgentScope, and LangGraph provide reusable abstractions for these fundamental primitives~\citep{wu2023autogen,hong2024metagpt,agentscope,langgraph}. Because customizing these frameworks beyond their built-in abstractions typically requires directly programming agent communication and orchestration logic, it can make development cumbersome and raise the barrier to rapid prototyping for users with limited programming experience~\citep{dibia2024autogenstudio,agentscope}.

To lower this development barrier, no-code and visual agent builders, including AutoGen Studio, AI2Apps, Dify, Langflow, and Flowise, offer reusable canvas components~\citep{dibia2024autogenstudio,ai2apps,dify,langflow,flowise}. However, agents remain steps in author-defined workflows: users must predefine control flow and loop boundaries and manually bind persistent outputs to later agent inputs~\citep{dify_loop,langflow_loop,flowise_agentflow}. This state plumbing becomes cumbersome when multiple feedback cycles interact, and some platforms further restrict nested loops~\citep{dify_nested_loop}, complicating MAS patterns with evolving contexts such as planner--executor and generator--reviewer--reviser loops~\citep{magenticone,chen2024mindsearch,qian2024chatdev,antoniades2024swe,li2025metal}.

We therefore present \textbf{ChatDev 2.0: DevAll} (hereafter DevAll), a no-code platform that combines an intuitive user interface for high-level interaction with a robust underlying engine as shown in Fig~\ref{fig:teaser}. Specifically, this engine integrates two core subsystems: (i) the MAS Compilation Engine, which compiles seven heterogeneous node types and semantic edges that decouple data flow from control flow, dynamically assembling agent contexts from runtime messages; and (ii) the MAS Execution Engine, which is powered by our proposed \textbf{CADET} (\textbf{C}ycle-\textbf{A}ware \textbf{D}ynamic \textbf{E}xecution \textbf{T}opology) algorithm—a cycle-aware scheduler that orchestrates multiple and nested feedback cycles while preserving iteration-level states, thereby enabling the correct and efficient execution of MAS graphs with arbitrary cycles. As an integrated whole, DevAll enables users to author, execute, and inspect dynamic MAS without writing any task-specific orchestration code.

To evaluate DevAll's expressiveness, we reconstruct representative systems for scientific data visualization, deep research report generation, and software development using only declarative graph specifications. Across the three scenarios, DevAll reproduces the core workflows of specialized systems and achieves competitive performance within a unified platform. This paper makes three contributions:
\begin{itemize}
    \item We open-source DevAll, a no-code platform for authoring, executing, and inspecting heterogeneous multi-agent systems.
    \item  We enable the execution of MAS graphs with arbitrary cycles via \textbf{CADET}, supporting rich and dynamic multi-agent interaction patterns.
    \item We validate DevAll across visualization, deep research, and software development, demonstrating that declarative graphs can reproduce representative specialized workflows with competitive performance.
\end{itemize}

\section{Related Work}

\paragraph{LLM-Based Multi-Agent Systems.}
LLM-based multi-agent systems coordinate specialized agents through role decomposition, message exchange, and feedback to solve complex tasks~\citep{chen2024agentverse,wu2023autogen}. General-purpose frameworks such as AutoGen~\citep{wu2023autogen} and AgentScope~\citep{agentscope} provide programmable abstractions for stateful agents, tools, messaging, and orchestration. AutoGen further offers GraphFlow for explicitly specified conditional and cyclic agent graphs and Studio for visual team configuration and testing~\citep{autogen_graphflow,dibia2024autogenstudio}. Other graph-based methods represent or optimize agent collaboration structures~\citep{gptswarm,gdesigner,qian2025scaling}. Despite their broad use in software engineering, deep research, and visualization, specifying and debugging multi-agent orchestration remains engineering-intensive~\citep{qian2024chatdev,prabhakar2025enterprisedeepresearch,chen2026coda,agentscope,wang2026teachmastergenerativeteachingcode}, motivating infrastructure that preserves MAS-level interaction semantics while reducing orchestration code.

\paragraph{No-Code and Visual Agent Builders.}
No-code and visual platforms expose agents, tools, retrieval components, and control operators on a canvas~\citep{dibia2024autogenstudio,dify,flowise,langflow,coze,openaiagentbuilder,copilotstudio}. Although these platforms increasingly support conversation memory and shared workflow state, their primary abstraction remains workflow-centric: the workflow owns control and state, while agents are invoked as steps whose inputs are assembled through node parameters, variable references, or prompt templates. Cyclic behavior is likewise expressed through explicit workflow constructs: Dify uses Loop and Iteration containers~\citep{dify_loop}, Langflow provides a list-oriented Loop component~\citep{langflow_loop}, and Flowise uses Iteration subflows and an explicit jump-back Loop node~\citep{flowise_agentflow}. Authors must therefore delineate loop scopes and explicitly route persistent outputs or shared variables across interacting feedback paths; some visual editors further restrict direct composition of nested loop constructs~\citep{dify_nested_loop}. DevAll instead adopts an agent-centric abstraction: each agent owns and evolves its local state, while interaction contexts are accumulated and propagated automatically along semantic edges. Cyclic and acyclic interactions use the same edge abstraction, so the visual graph directly corresponds to the MAS being executed, without special loop nodes.

\section{System Design}
\label{sec:engines}

\begin{figure*}[t]
\centering
\includegraphics[width=\textwidth]{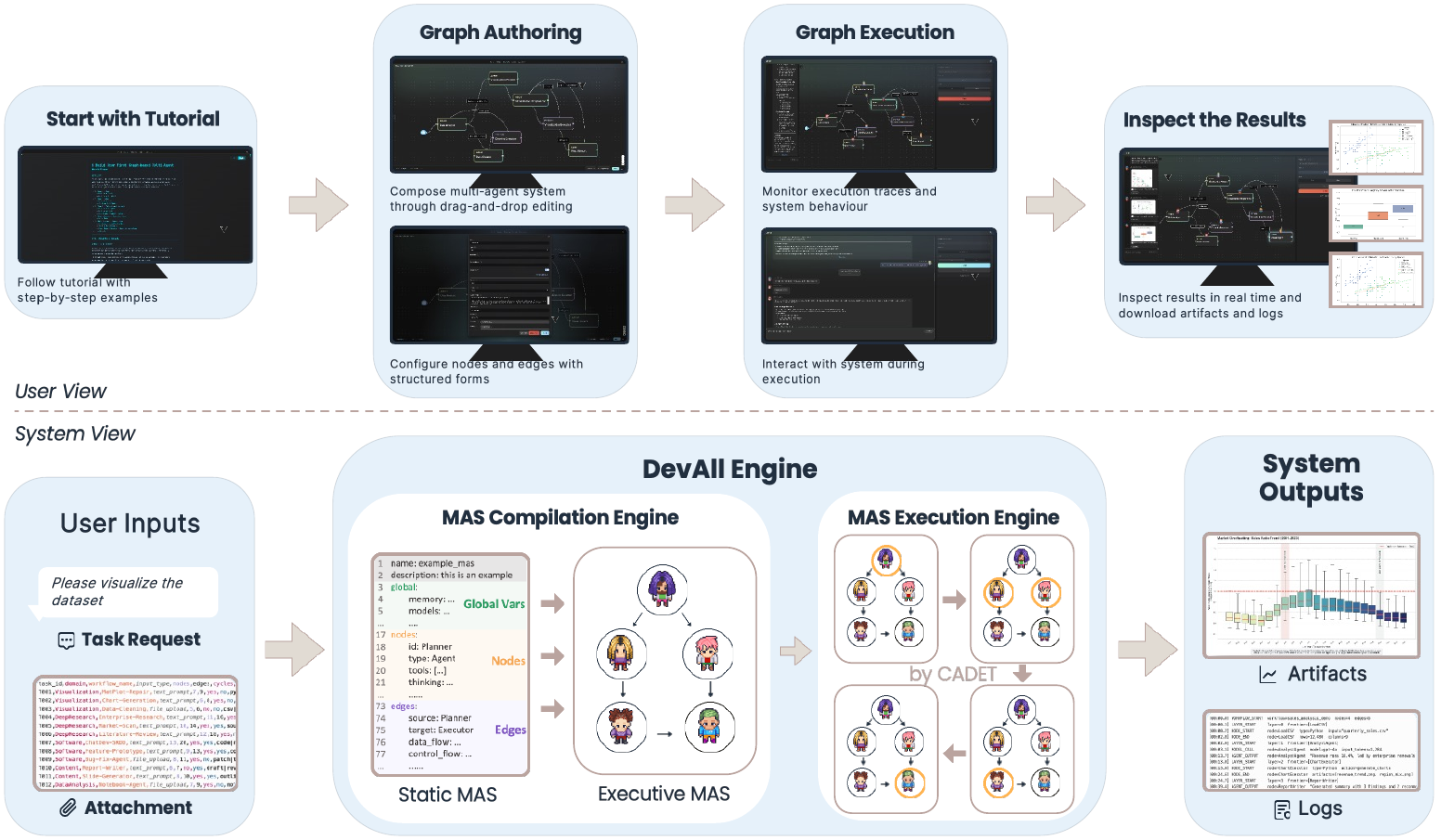}
\caption{Overview of DevAll. The upper part shows the user interface for graph authoring, execution launching, and result inspection; the lower part presents the underlying engine, which includes the MAS Compilation Engine that compiles declarative specifications into executable graphs, and the CADET runtime that schedules cycle-aware execution over semantic edges to produce final results along with intermediate traces and artifacts.}
\label{fig:system-overview}
\end{figure*}

DevAll is a no-code platform for developing heterogeneous artifacts across scenarios like data visualization, software engineering, and deep research. Its two core components are the \emph{MAS Compilation Engine}, which turns declarative specifications into executable graphs, and the cycle-aware \emph{MAS Execution Engine}, which schedules and runs those graphs. Together they decouple authoring from runtime: users configure agents, tools, and dependencies on a visual canvas, while the engine handles instantiation, scheduling, and monitoring. When a request runs, DevAll binds inputs, routes information through connected nodes, and surfaces outputs and traces for inspection and optional human feedback.

\subsection{MAS Compilation Engine}
\label{sec:graph-builder}

The first engine, the \emph{MAS Compilation Engine}, compiles a human-readable YAML specification~\citep{aws_yaml_json}, which serves as the canonical on-disk representation for distributing and sharing MAS graph designs, into an executable directed graph~\citep{agentscope, graphofagents}, which provides a uniform abstraction over heterogeneous multi-agent systems.
\[
\mathcal{G} = (\mathcal{V}, \mathcal{E}, \Theta)
\]
 The specification declares the global configuration $\Theta$, which holds graph-level runtime settings such as shared memory, model and tool registries, workspace, and permissions, together with the nodes and edges that the compilation engine instantiates below.

The node set $\mathcal{V}$ defines the executable units of the MAS. During graph compilation, the engine instantiates each node through a registry that maps node types to configuration schemas and executors, enabling a unified interface while preserving type-specific semantics. The seven built-in node types in table~\ref{tab:node-types} capture the common operations observed across practical MAS workloads.

The edge set $\mathcal{E}$ carries the abstraction that distinguishes DevAll. Unlike an edge that merely forwards context, a \emph{semantic edge} specifies both how information is passed and how downstream execution is activated, expressed as a tuple $e=(u,v,c_e,\phi_e,\psi_e)$, where $\operatorname{src}(e)=u$ and $\operatorname{dst}(e)=v$ denote the source and target nodes, respectively. The activation condition $c_e$ checks whether the edge should fire after the source $u$ emits an output; the data-flow policy $\phi_e$ prescribes how that output is merged, transformed, retained, or cleared in the target context of $v$; and the control-flow policy $\psi_e$ governs how $v$'s scheduling state is updated. By decoupling data movement from scheduling triggers, DevAll can route information and orchestrate execution independently, enabling richer MAS semantics than fixed workflows.
MAS compilation engine resolves each edge's endpoints and attaches these policies from the specification, turning the edge into an executable communication rule rather than a static link.

After instantiating all nodes and edges, the compilation engine validates graph consistency, and passes the resulting graph $\mathcal{G}$ to the MAS Execution Engine.

\subsection{MAS Execution Engine}
\label{sec:mas-execution}

\begin{figure*}[t]
    \centering
    \includegraphics[width=\textwidth]{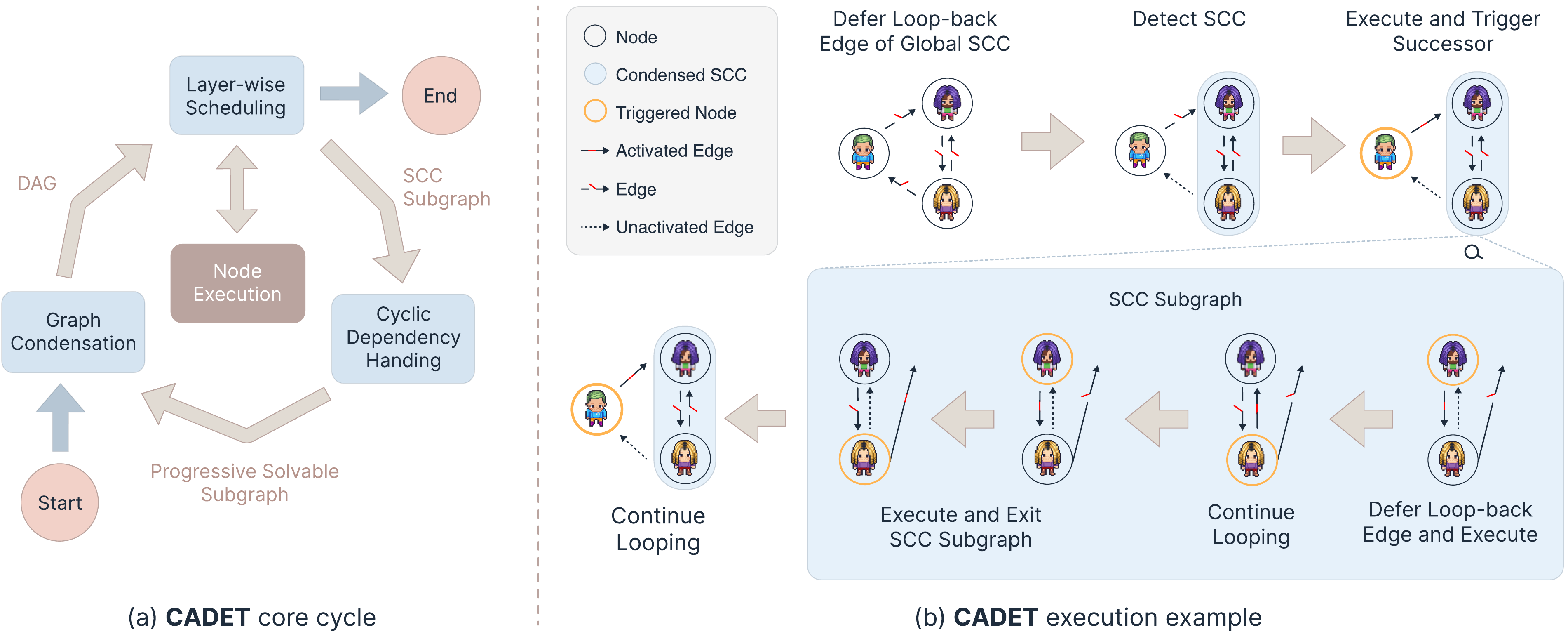}
    \caption{CADET execution over MAS graphs. CADET contracts cyclic regions into condensed nodes for layer-wise scheduling, and executes the corresponding SCC subgraphs through a scoped iterative process.}
    \label{fig:cadet-execution}
\end{figure*}
The MAS Execution Engine interprets $\mathcal{G}$ through two primitives: \emph{node execution} and \emph{semantic edge propagation}. A node becomes eligible to run when it is triggered by the initial user request or by an activated incoming edge; its executor consumes the local context, performs node-specific computation, generates artifacts, and emits messages. The runtime then evaluates the node's outgoing semantic edges according to Section~\ref{sec:graph-builder}: each active edge applies its data-flow policy $\phi_e$ to the target context and its control-flow policy $\psi_e$ to the target execution state. Nodes whose readiness conditions become satisfied join the execution frontier. For an acyclic graph, these primitives advance the frontier along a topological order, yielding a deterministic layer-wise schedule. This is the baseline execution model we preserve.

MAS graphs may contain cycles, for which no topological order exists. Consequently, the baseline layer-wise schedule cannot be applied directly. We therefore introduce CADET to address this challenge by condensing cyclic regions into executable strongly connected components (SCCs) subgraphs, yielding an acyclic condensation graph for global scheduling while handling each SCC through scoped iterative execution, as illustrated in Figure~\ref{fig:cadet-execution}.

\paragraph{Graph condensation}
CADET decomposes $\mathcal{G}$ into SCCs $\{\mathcal{C}_1, \ldots, \mathcal{C}_K\}$~\citep{tarjan1972dfs}. Singleton SCCs without self-loops remain ordinary nodes, while cyclic SCCs, including multi-node SCCs and singleton self-loop components, are contracted into \emph{condensed nodes}~\citep{bangjensen2009digraphs}. Let
\[
\begin{aligned}
\mathcal{V}^\dagger
&= \{\mathcal{C}_1,\ldots,\mathcal{C}_K\},\\
\mathcal{E}^\dagger
&= \bigl\{(\mathcal{C}_i,\mathcal{C}_j)
\ \bigm|\ i\neq j,\ \exists e\in\mathcal{E}:\\
&\qquad
\operatorname{src}(e)\in\mathcal{C}_i,\quad
\operatorname{dst}(e)\in\mathcal{C}_j
\bigr\}.
\end{aligned}
\]
and define the condensation graph as $\mathcal{G}^\dagger = (\mathcal{V}^\dagger, \mathcal{E}^\dagger)$. Because edges only connect different components, $\mathcal{G}^\dagger$ is a DAG and therefore schedulable like the baseline; each condensed node retains its internal subgraph $\mathcal{C}_k$ for iterative execution.

\paragraph{Cyclic dependency handling}
Inside a condensed SCC $\mathcal{C}_k$, CADET executes it as a scoped iterative subgraph with internal edges $\mathcal{E}_k=\{(u,v)\in\mathcal{E}\mid u,v\in\mathcal{C}_k\}$. Starting from the unique active entry of $\mathcal{C}_k$, CADET identifies $\mathcal{B}_k$ recursively: at each SCC scope, it marks the internal edges targeting that scope's active entry as back edges and applies the same rule to any cyclic SCC remaining after those edges are removed, partitioning the internal edges as
\[
\mathcal{E}_k = \widehat{\mathcal{E}}_k \cup \mathcal{B}_k, \qquad
\widehat{\mathcal{E}}_k \cap \mathcal{B}_k = \emptyset .
\]
The non-back edges $\widehat{\mathcal{E}}_k$ form an acyclic skeleton and execute within the current iteration, while back edges $\mathcal{B}_k$ are treated as loop-carried dependencies whose outputs are deferred to the next iteration. Each iteration activates nodes triggered by incoming external edges or back edges from the previous round. If an active edge exits $\mathcal{C}_k$, CADET propagates its output to the downstream node in $\mathcal{G}^\dagger$; otherwise, if back edges remain deferred, it proceeds to iteration $t+1$ with updated contexts and trigger states. A configurable iteration limit bounds this process and ensures termination.


\paragraph{Layer-wise scheduling}
On the acyclic scheduling view $\mathcal{G}^\dagger$, CADET first computes topological layers
\[
\mathcal{L}^\dagger =
(\mathcal{L}_0^\dagger,\ldots,\mathcal{L}_M^\dagger)
= \mathrm{TopoLayers}(\mathcal{G}^\dagger).
\]
The user request is bound to the configured start nodes, which initialize the trigger states for the first scan. CADET then scans the layers in order. At layer $t$, an item $x\in\mathcal{L}_t^\dagger$ is executed only when $\mathrm{Ready}_t(x)$ holds, where $\mathrm{Ready}_t$ captures both topological dependency order and runtime trigger states induced by the user request and activated semantic edges. An ordinary node consumes its current trigger state, executes on its local context, and propagates outputs through activated outgoing edges. When CADET encounters a condensed node $\mathcal{C}_k$, it expands the node and delegates execution to the cyclic dependency handling described above.

\subsection{System Interface}
\label{sec:system-interface}
Figure~\ref{fig:system-overview} shows DevAll's integrated no-code interface, which serves as the user-facing layer over the MAS Compilation Engine and MAS Execution Engine. The interface organizes MAS development into four connected views: Tutorial, Authoring, Execution, Inspection. Together, these views support example-based onboarding, visual graph construction, interactive execution, and batch-level experimentation, while keeping both the graph structure and runtime behavior inspectable.

\paragraph{Graph authoring}
The Tutorial view provides example MAS graphs that users can inspect and reuse as starting points. In the authoring view, users construct a MAS by placing heterogeneous node types on a canvas and connecting them with semantic edges. Configuration panels are generated from metadata exported by the MAS Compilation Engine. Each node type, edge processor, memory store, and related option declares its own configurable fields, and the interface renders the relevant fields recursively after a concrete type or option is selected. This metadata-driven design improves system extensibility while preserving the no-code user experience: users edit MAS components through structured controls, and developers can add new component types by registering their configuration metadata rather than designing a separate interface panel.

\paragraph{Interactive execution}
The Launch view binds a user request and optional input files to the selected executable graph, then passes them to the configured start nodes. During execution, it receives runtime events from the MAS Execution Engine and displays node status, intermediate messages, and generated artifacts as they are produced. If execution reaches a Human node, DevAll pauses the corresponding path, collects feedback through the same interface, and resumes downstream execution after the feedback is submitted. The completed run remains inspectable through its node-level traces, execution logs, token-usage records, and workspace artifacts. Users can download execution records for offline inspection, sharing, or subsequent reuse.

\paragraph{Experiments}
For repeated runs and batch experiments, DevAll provides a Laboratory interface together with a Python SDK for programmatic execution and large-scale evaluation. This separation keeps the launch-time interface lightweight while still supporting systematic inspection and evaluation when users need it.

\section{Experiment}
\begin{table*}[htbp]
\centering
\small
\setlength{\tabcolsep}{7pt}
\renewcommand{\arraystretch}{1.12}

\caption{Detailed metric breakdown across benchmarks. For MatPlotBench, M1--M3 denote execution pass rate, code quality score, and visualization success rate; for DeepResearchBench, M1--M4 denote comprehensiveness, insight, instruction-following, and readability; for SRDD, M1-M3 denote completeness, executability, and consistency, respectively. Overall denotes the benchmark-specific aggregate score. $\Delta$ denotes DevAll $-$ Reference, and higher values indicate better performance. Detailed metric definitions are provided in the appendix~\ref{app:evaluation-details}.}
\label{tab:main-results}
\begin{tabular}{@{}llccccc@{}}
\toprule
\textbf{Benchmark} & \textbf{Method}
& \textbf{M1} & \textbf{M2} & \textbf{M3} & \textbf{M4} & \textbf{Overall} \\
\midrule

\multirow{3}{*}{\shortstack[l]{MatPlotBench}}
& CoDA & 0.8300 & 0.8750 & 0.6730 & -- & 0.7130 \\
& DevAll & \textbf{0.9900} & \textbf{0.8950} & \textbf{0.7160} & -- & \textbf{0.7950} \\
\rowcolor{DeltaRow}
& $\Delta$ & {\scriptsize\textcolor{gray}{+0.1600}} & {\scriptsize\textcolor{gray}{+0.0200}} & {\scriptsize\textcolor{gray}{+0.0430}} & -- & {\scriptsize\textcolor{gray}{+0.0820}} \\
\addlinespace[0.35em]

\multirow{3}{*}{\shortstack[l]{DeepResearchBench}}
& EDR & \textbf{0.3230} & \textbf{0.3274} & \textbf{0.3700} & \textbf{0.4192} & \textbf{0.3500} \\
& DevAll & 0.3048 & 0.3132 & 0.3390 & 0.4147 & 0.3319 \\
\rowcolor{DeltaRow}
& $\Delta$ & {\scriptsize\textcolor{gray}{-0.0182}} & {\scriptsize\textcolor{gray}{-0.0142}} & {\scriptsize\textcolor{gray}{-0.0310}} & {\scriptsize\textcolor{gray}{-0.0045}} & {\scriptsize\textcolor{gray}{-0.0181}} \\
\addlinespace[0.35em]

\multirow{3}{*}{\shortstack[l]{SRDD}}
& ChatDev 1.0 & 0.9670 & \textbf{0.8380} &  \textbf{0.7916} & -- &\textbf{0.6574} \\
& DevAll & \textbf{0.9830} & 0.8250 &  0.7762 &-- & 0.6509 \\
\rowcolor{DeltaRow}
& $\Delta$ & {\scriptsize\textcolor{gray}{+0.0160}} & {\scriptsize\textcolor{gray}{-0.0130}} & {\scriptsize\textcolor{gray}{-0.0154}} &-- & {\scriptsize\textcolor{gray}{-0.0065}} \\
\bottomrule
\end{tabular}
\end{table*}
\subsection{Experimental Setup}
The goal of our evaluation is not to surpass domain-specific systems, but to test whether a single no-code platform can reproduce the core workflow behavior of specialized MAS across heterogeneous domains. We therefore select three representative settings with substantially different output modalities: scientific data visualization on MatPlotBench~\citep{yang-etal-2024-matplotagent}, deep research report generation on DeepResearchBench~\citep{du2025deepresearch}, and software generation on SRDD~\citep{qian2024chatdev}. These settings correspond to three workflow-based reference systems: CoDA~\citep{chen2026coda}, Enterprise Deep Research~\citep{prabhakar2025enterprisedeepresearch}, and ChatDev 1.0~\citep{qian2024chatdev}, respectively.

For each setting, we re-express the original workflow as a declarative MAS graph in DevAll, \emph{without introducing any task-specific orchestration code}; users author only YAML specifications of nodes and semantic edges through the platform. To isolate workflow reproduction from confounding factors, we standardize the backbone model as GPT-4o and keep tool configurations consistent across comparisons wherever possible, so that performance differences primarily reflect the fidelity of the reproduced workflow rather than variations in the underlying model or tools. In the DeepResearchBench setting, we use Serper as the search backend and GPT-5.5 as the LLM judge; in the SRDD setting, we use \texttt{text-embedding-ada-002} as the embedding model. Detailed metric definitions are provided in Appendix~\ref{app:evaluation-details}.

\subsection{Cross-Domain Results}
Table~\ref{tab:main-results} reports results across visualization, deep research, and software generation. The main observation is \emph{generality}: the same no-code platform can reproduce three specialized workflow systems and attain competitive performance with each domain-specific implementation, without relying on task-specific orchestration code.

On MatPlotBench, the DevAll reproduction matches and slightly exceeds CoDA, with the overall score moving from 0.7130 to 0.7950. The gain is concentrated in execution pass rate (0.8300 to 0.9900), which is consistent with DevAll's unified runtime providing stable workspace management and error handling rather than changing the workflow logic itself. On DeepResearchBench, the reproduced workflow yields an overall score of 0.3319 versus 0.3500. Given that DeepResearchBench relies entirely on LLM-as-judge, which exhibits non-negligible variance, this small difference is within the expected range of evaluation noise. On SRDD, the reproduced workflow achieves higher completeness but slightly lower executability and consistency, with the overall score remaining nearly identical.

Taken together, these results show that DevAll can host structurally and behaviorally distinct multi-agent workflows within one unified platform while preserving task effectiveness. This supports our central claim that a multi-agent system can be developed, executed, and reproduced as a no-code executable object across diverse domains.

\subsection{Workflow Construction and Runtime Overhead}
We evaluate the overhead introduced by DevAll at two stages of the workflow
lifecycle. Construction overhead captures the one-time effort required to
represent and configure a workflow, quantified by specification size and
estimated canvas-level operations. Runtime overhead captures the additional
latency introduced by CADET during graph compilation, planning, and local
scheduling, excluding network requests and LLM inference.
\begin{table}[t]
\centering
\scriptsize
\setlength{\tabcolsep}{4pt}
\renewcommand{\arraystretch}{1.08}
\caption{Structural construction complexity of the reproduced workflows.
Est. ops. is the sum of nodes, edges, and configuration items and serves as a
proxy for canvas-level configuration operations.}
\label{tab:workflow-complexity}
\begin{tabular}{@{}lrrrr@{}}
\toprule
\textbf{Method}
& \textbf{Nodes}
& \textbf{Edges}
& \shortstack{\textbf{Config.}\\\textbf{items}}
& \shortstack{\textbf{Est.}\\\textbf{ops.}} \\
\midrule
CoDA        & 14 & 26 & 88  & 128 \\
EDR         & 8  & 11 & 65  & 84  \\
ChatDev 1.0 & 26 & 54 & 178 & 258 \\
\bottomrule
\end{tabular}
\end{table}

We quantify workflow construction overhead using three structural metrics:
\textit{Nodes}, \textit{Edges}, and \textit{Config. items}. Nodes denote
workflow components, Edges denote directed connections, and Config. items
denote non-empty semantic node settings or non-default edge conditions. We
define estimated operations (Est. ops.) as
$N_{\mathrm{node}}+N_{\mathrm{edge}}+N_{\mathrm{config}}$.
As shown in Table~\ref{tab:workflow-complexity}, reproducing the three workflows required 84--258 estimated structural operations. In DevAll, these operations
typically correspond to point-and-click interactions for placing nodes, connecting components, and selecting configuration options. Est. ops. therefore
approximates structural configuration burden rather than the exact number of mouse clicks or keystrokes. Additional comparisons of YAML and reference-source
sizes are reported in Appendix Table~\ref{tab:appendix-workflow-compactness}. Across the evaluated workflows,
construction burden remained below 300 estimated operations, indicating that structurally diverse multi-agent workflows can be configured with a controlled
amount of canvas-level interaction.

We next measure the additional runtime overhead introduced by CADET. We scale
the number of cyclic SCCs from 1 to 16 and separately measure compilation,
planning, and execution. Compilation covers YAML loading, validation,
executable-graph construction, and planning. The planning benchmark isolates
SCC detection, graph condensation, and topological layering on a preconstructed
graph. Execution measures the local scheduling path without network requests or LLM inference.

As shown in Table~\ref{tab:cadet_overhead}, median compilation, planning, and
execution times increased approximately linearly over the evaluated graph
sizes. At the largest configuration, median compilation and execution required
16.23 and 14.46~ms, respectively, while planning required 91.96~$\mu$s.
These results indicate that CADET adds modest local processing latency compared
with the latency typically incurred by model inference.

\begin{table}[!htbp]
\centering
\caption{Compilation and CADET overhead scaling. Times are p50/p95 over
1,000 runs after five warm-ups; compilation includes planning.}
\label{tab:cadet_overhead}
\scriptsize
\setlength{\tabcolsep}{2.3pt}
\begin{tabular}{@{}rcccc@{}}
\toprule
SCCs & $|\mathcal{V}|/|\mathcal{E}|$
& Compile (ms) & Plan ($\mu$s) & Execute (ms) \\
\midrule
1  & 6/7    & 1.53/1.60  & 7.13/7.25   & 1.19/1.33 \\
2  & 10/14  & 2.51/2.64  & 11.54/11.79 & 2.10/2.49 \\
4  & 18/28  & 4.43/4.60  & 21.38/21.75 & 3.88/4.26 \\
8  & 34/56  & 8.35/8.81  & 42.29/42.88 & 7.27/7.71 \\
16 & 66/112 & 16.23/35.36 & 91.96/93.04 & 14.46/15.62 \\
\bottomrule
\end{tabular}
\end{table}

Together, these results show that DevAll balances workflow expressiveness with engineering efficiency. Across the evaluated settings, its unified abstraction supported structurally diverse multi-agent workflows while maintaining moderate construction burden and modest runtime orchestration overhead.

\section{Conclusion}
We presented \textbf{DevAll}, a no-code platform that treats a multi-agent system, rather than a static workflow, as the executable object. DevAll compiles a declarative YAML specification into an executable graph whose \emph{semantic edges} decouple data flow from control flow, and runs it with \textbf{CADET}, a scheduling protocol that restores deterministic layer-wise execution to cyclic MAS graphs by condensing strongly connected components into an acyclic topology. An integrated interface lets users author, launch, monitor, and inspect these graphs without writing code, with support for human-in-the-loop intervention and trace inspection. Across visualization, deep research, and software generation, DevAll reproduces three domain-specific workflow systems within a single unified platform, attaining competitive performance without task-specific orchestration code. These results demonstrate that dynamic, feedback-driven multi-agent systems can be built, executed, and inspected as no-code executable objects. Future work includes extending the scheduling algorithm with more sophisticated loop control mechanisms, and building a wider range of representative MAS on top of DevAll to serve diverse application domains including scientific discovery, automated data analysis, and interactive education.

\newpage
\section*{Limitations}
Although DevAll reduces implementation effort, users must still translate task
requirements into workflow structures and manually specify agent roles, prompts,
dependencies, and control conditions. Thus, no-code authoring simplifies
implementation but does not yet automate workflow design. Moreover, the current
graph abstraction and component ecosystem may not cover every domain-specific
interaction pattern, and specialized integrations may require custom extensions.
As workflows grow, visual organization, modular maintenance, version evolution,
and the transfer of accumulated design experience also offer opportunities for
improvement. Future work will explore intelligent authoring assistance, reusable
workflow templates, stronger lifecycle management, and user studies that assess
authoring efficiency and learnability across different levels of expertise.

\section*{Acknowledgments}
This work was supported by the Tsinghua University (Department of Computer Science and Technology)-Siemens Ltd., China Joint Research Center for Industrial Intelligence and Internet of Things (JCIIOT), and by the Shanghai Municipal Special Program for Basic Research on General AI Foundation Models (Grant No. 2025SHZDZX026D04), in collaboration with the Shanghai Artificial Intelligence Laboratory. And we thank Fu Fangyu for his contributions to the frontend development of the DevAll platform.
\bibliography{custom}
\clearpage
\newpage
\newpage
\appendix
\section{Implementation Details}
\label{app:implementation}

We provide omitted implementation details: built-in node types and CADET pseudocode.

\subsection{Built-in Node Types}
\label{app:node-types}

Table~\ref{tab:node-types} summarizes the built-in nodes used to construct executable MAS graphs.

\begin{table}[ht]
\centering
\small
\setlength{\tabcolsep}{4pt}
\caption{Built-in node types in DevAll.}
\label{tab:node-types}
\renewcommand{\arraystretch}{1.15}
\begin{tabular}{@{}p{0.22\columnwidth}p{0.70\columnwidth}@{}}
\toprule
\textbf{Node type} & \textbf{Role} \\
\midrule
Agent & LLM-backed agent configured with prompt, tools, memory, and an optional thinking module. \\
Human & Human-feedback checkpoint that pauses execution until input is submitted. \\
Python & Workspace Python executor that returns standard output. \\
Subgraph & Nested MAS graph defined inline or loaded from YAML. \\
Literal & Static message emitter. \\
Passthrough & Context forwarder that leaves inputs unchanged. \\
LoopCounter & Iteration gate that releases output after configured rounds. \\
\bottomrule
\end{tabular}
\end{table}

\subsection{CADET Execution Algorithm}
\label{app:cadet-execution}

CADET executes MAS graphs by scheduling the condensation graph and resolving cycles inside condensed components. Algorithms~\ref{alg:app-cadet-execution} and~\ref{alg:app-cadet-scc} give the two procedures. $\textsc{SemProp}$ applies $\phi_e$ to target context and $\psi_e$ to target state; $\textsc{Run}$ returns activated edges.

\begin{algorithm}[ht]
\caption{CADET Execution over the Condensation Graph}
\label{alg:app-cadet-execution}
\begin{algorithmic}[1]
\Require Executable MAS graph $\mathcal{G}=(\mathcal{V},\mathcal{E},\Theta)$, start nodes $\mathcal{S}$, runtime context $\Gamma$
\State $\mathcal{G}^{\dagger}=(\mathcal{V}^{\dagger},\mathcal{E}^{\dagger}) \gets \textsc{Condense}(\textsc{SCCDecompose}(\mathcal{G}))$
\State $\mathcal{F} \gets \textsc{InitFrontier}(\mathcal{S},\Gamma)$; $S \gets \emptyset$
\While{$\mathcal{F} \neq \emptyset$}
    \State Execute each $x \in \mathcal{F}$ in parallel when possible
    \State For ordinary nodes, run $\textsc{ExecuteNode}(x,\Gamma)$ and $\textsc{SemProp}(x,\Gamma)$; for SCCs, run $\textsc{ExecuteSCC}(x,\Gamma)$
    \State $S \gets S \cup \mathcal{F}$; $\mathcal{F} \gets \{x \in \mathcal{V}^{\dagger}\setminus S \mid \textsc{Ready}(x)\}$
\EndWhile
\end{algorithmic}
\end{algorithm}

\begin{algorithm}[!t]
\caption{Cyclic Dependency Handling inside a Condensed SCC}
\label{alg:app-cadet-scc}
\begin{algorithmic}[1]
\Require Condensed SCC $\mathcal{C}_k$, internal edges $\mathcal{E}_k$, runtime context $\Gamma$
\State $\mathcal{B}_k \gets \textsc{LoopBackEdges}(\mathcal{C}_k,\mathcal{E}_k)$; $\widehat{\mathcal{E}}_k \gets \mathcal{E}_k \setminus \mathcal{B}_k$
\State $Q \gets \textsc{InitTriggers}(\mathcal{C}_k)$; $t \gets 0$
\Repeat
    \State $\textsc{ApplyDeferred}(Q,\Gamma)$
    \State $\begin{aligned}
    \mathcal{L} \gets {} & \textsc{TopologicalLayers}( \\
    & \textsc{BuildDAG}(\mathcal{C}_k,\widehat{\mathcal{E}}_k,Q))
\end{aligned}$
    \State $exit \gets \textsc{false}$; $Q' \gets \emptyset$
    \For{layer $L \in \mathcal{L}$}
        \For{triggered $x \in L$ in parallel when possible}
        \For{activated edge $e=(u,w) \in \textsc{Run}(x,\Gamma)$}
            \If{$w \notin \mathcal{C}_k$}
                \State $\textsc{SemProp}(e,\Gamma)$; $exit \gets \textsc{true}$
            \ElsIf{$e \in \mathcal{B}_k$}
                \State $Q' \gets Q' \cup \{e\}$
            \Else
                \State $\textsc{SemProp}(e,\Gamma)$
            \EndIf
        \EndFor
        \EndFor
    \EndFor
    \State $Q \gets Q'$; $t \gets t+1$
\Until{$exit$ or $Q=\emptyset$ or a loop limit is reached}
\end{algorithmic}
\end{algorithm}

\section{Evaluation Details}
\label{app:evaluation-details}
We follow the evaluation protocols used in the original MatPlotBench, DeepResearchBench, and SRDD benchmark papers and summarize the corresponding metrics below.

\subsection{MatPlotBench Metrics}
\label{app:matplotbench-metrics}

MatPlotBench evaluates code execution and chart quality using EPR, CS, VSR, and OS.

\paragraph{Execution Pass Rate (EPR).}
$\text{EPR}=|\{q \in Q : \mathrm{exec}(c_q)=\mathrm{success}\}|/|Q|$, where $c_q$ denotes the generated code for query $q$.

\paragraph{Code Quality Score (CS).}
$\text{CS}=\frac{1}{|Q|}\sum_{q \in Q}s_c(q)$, where $s_c(q)$ is the LLM-evaluated code score on a 0--100 scale.

\paragraph{Visualization Success Rate (VSR).}
$\text{VSR}=\frac{1}{|Q_{\mathrm{exec}}|}\sum_{q \in Q_{\mathrm{exec}}}s_v(q)$, where $s_v(q)$ is the vision-based score and $Q_{\mathrm{exec}}$ contains executable cases.

\paragraph{Overall Score (OS).}
$\text{OS}=\frac{1}{|Q|}\sum_{q \in Q}(s_c(q)+s_v(q))/2$, where $s_v(q)=0$ for non-executable cases.

\subsection{DeepResearchBench Evaluation Procedure}
\label{app:deepresearchbench-eval}

DeepResearchBench evaluates reports with RACE: Comprehensiveness (COMP), Insight/Depth (DEPTH), Instruction-Following (INST), and Readability (READ).

\paragraph{Dynamic Weight Estimation.}
The judge LLM estimates task-specific dimension weights $w_d$, with $\sum_d w_d=1$.

\paragraph{Criteria Generation.}
The judge LLM generates criteria and weights for each dimension.

\paragraph{Reference-Based Pairwise Scoring.}
RACE scores the target and reference reports under the same criteria. It aggregates dimension scores as $S_{\mathrm{int}}(R)=\sum_d s_d \cdot w_d$ and normalizes by $S_{\mathrm{final}}(R_{\mathrm{tgt}})=S_{\mathrm{int}}(R_{\mathrm{tgt}})/(S_{\mathrm{int}}(R_{\mathrm{tgt}})+S_{\mathrm{int}}(R_{\mathrm{ref}}))$; 0.5000 indicates parity.

\paragraph{Evaluation Dimensions.}
COMP, DEPTH, INST, and READ measure coverage, analysis, requirement adherence, and organization.

\subsection{SRDD Metrics}
\label{app:srdd-metrics}

SRDD evaluates software projects using Completeness (C), Executability (E), Consistency (CS), and Quality (Q).

\paragraph{Completeness (C).}
C is the percentage of generated projects without placeholder code.

\paragraph{Executability (E).}
E is the percentage of generated projects that compile and run successfully.

\paragraph{Consistency (CS).}
CS is the embedding cosine similarity between the requirement and generated code.

\paragraph{Quality (Q).}
Q integrates the three metrics as $Q=C \times E \times CS$.

\section{Supplementary Results and Case Studies}
\label{app:supplementary-results}

We report workflow specification cost and representative generated artifacts.

\subsection{Workflow Specification Cost}
\label{app:workflow-specification-cost}

We estimate the authoring effort required to reproduce each specialized workflow. In DevAll, the YAML graph specification records the canvas configuration produced by user drag-and-drop and parameter-setting operations; a smaller YAML specification therefore indicates fewer user-side configuration actions. For each reference system, we count the task-specific orchestration and configuration code needed to define agents, prompts, control flow, state handling, and message passing. Reusable libraries, model APIs, evaluation scripts, data files, environment files, and generated outputs are excluded.

\definecolor{RatioLightYellow}{HTML}{FFF7DD}
\definecolor{RowWhite}{HTML}{FFFFFF}

\begin{table}[t]
\centering
\scriptsize
\setlength{\tabcolsep}{3.5pt}
\renewcommand{\arraystretch}{1.08}
\caption{Size of DevAll workflow specifications relative to reference workflow source code.}
\label{tab:appendix-workflow-compactness}
\resizebox{\columnwidth}{!}{
\begin{tabular}{@{} l r r r r r r @{}}
\toprule
\textbf{Method}
& \multicolumn{3}{c}{\textbf{Lines}}
& \multicolumn{3}{c}{\textbf{Characters}} \\
\cmidrule(lr){2-4}
\cmidrule(l){5-7}
& \textbf{YAML}
& \textbf{Source}
& \textbf{YAML/Source}
& \textbf{YAML}
& \textbf{Source}
& \textbf{YAML/Source} \\
\midrule
\rowcolor{RowWhite}
CoDA
& 928
& 10,276
& \cellcolor{RatioLightYellow}9.0\%
& 34,406
& 445,046
& \cellcolor{RatioLightYellow}7.7\% \\
\rowcolor{RowWhite}
EDR
& 664
& 14,286
& \cellcolor{RatioLightYellow}4.6\%
& 28,910
& 659,760
& \cellcolor{RatioLightYellow}4.4\% \\
\rowcolor{RowWhite}
ChatDev 1.0
& 1,042
& 1,916
& \cellcolor{RatioLightYellow}54.4\%
& 39,253
& 90,503
& \cellcolor{RatioLightYellow}43.4\% \\
\bottomrule
\end{tabular}
}
\end{table}

\noindent
Table~\ref{tab:appendix-workflow-compactness} reports both line and character counts. YAML/Source is the DevAll specification size divided by the corresponding reference source size; lower ratios indicate fewer canvas-level configuration operations relative to task-specific code implementation.

\subsection{Qualitative Case Studies}
\label{app:qualitative-examples}

\begin{figure}[!t]
\centering
\vspace{-0.6em}
\captionsetup{font=scriptsize}
\setlength{\tabcolsep}{1.5pt}
\begin{tabular}{@{}cc@{}}
\textbf{\scriptsize CoDA} & \textbf{\scriptsize DevAll} \\[-0.2em]
\includegraphics[width=0.40\columnwidth]{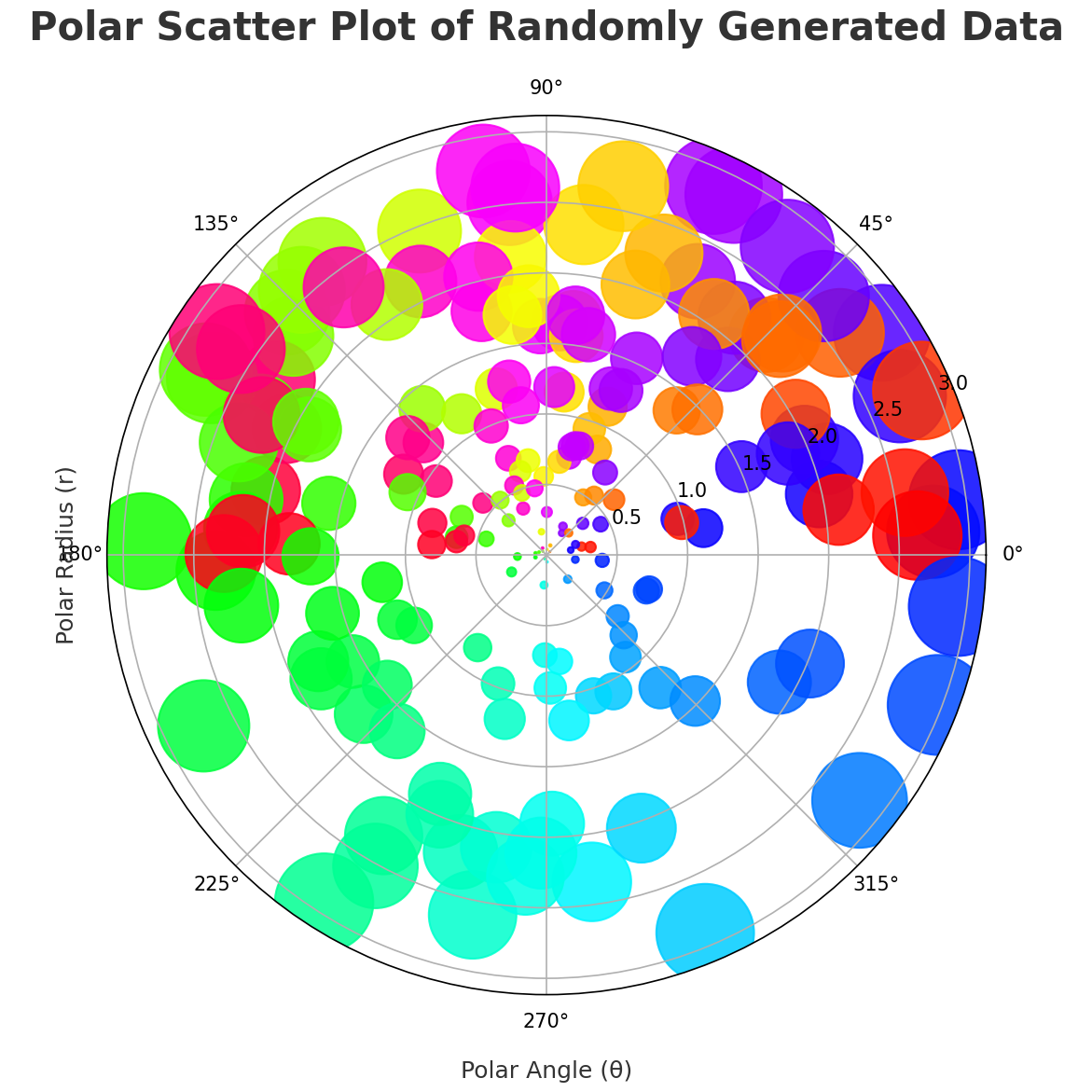}
&
\includegraphics[width=0.40\columnwidth]{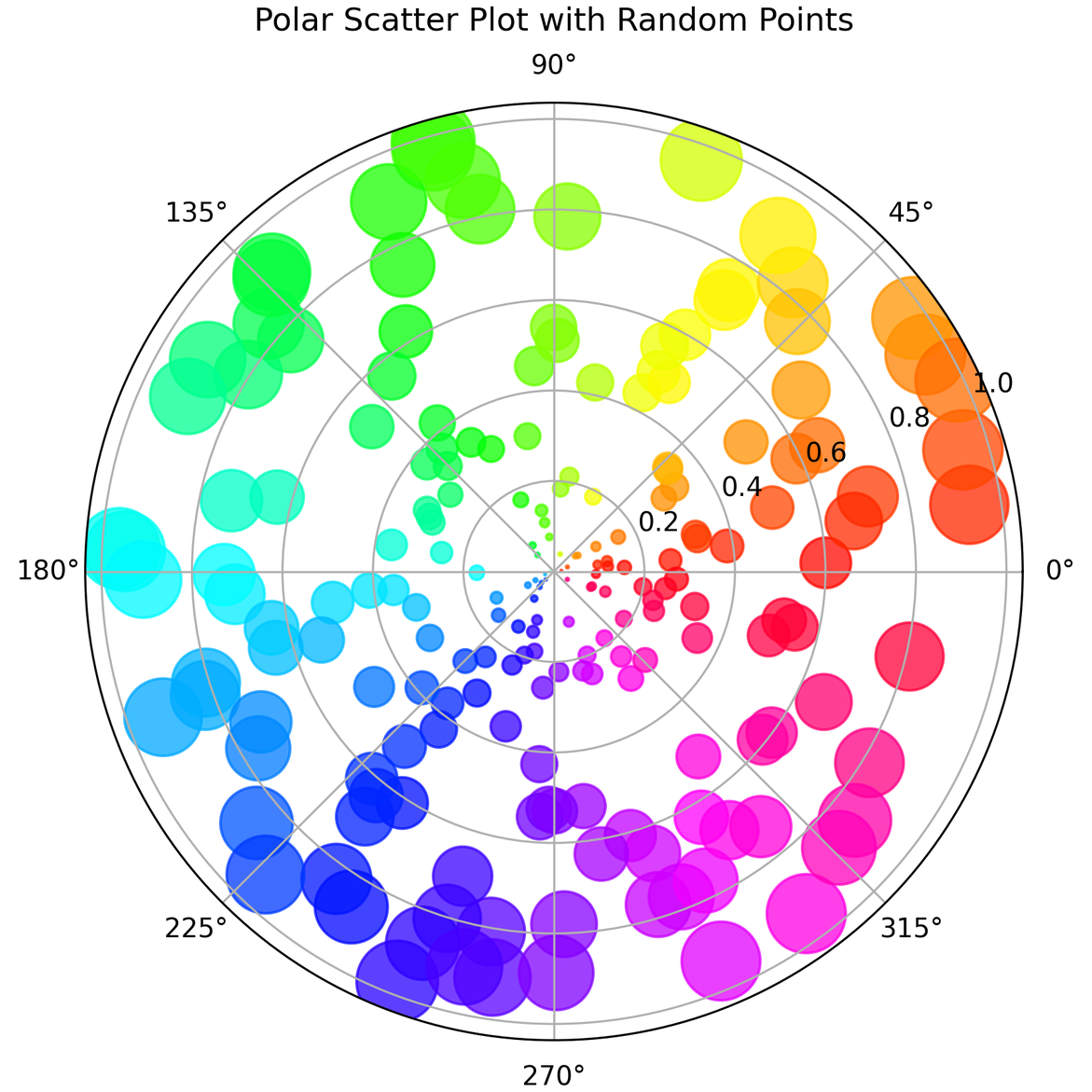}
\end{tabular}
\vspace{-0.8em}
\caption{Qualitative comparison of generated visualization artifacts on a representative MatPlotBench case.}
\label{fig:app-visualization-comparison}
\end{figure}

\begin{figure}[!t]
\centering
\vspace{-0.6em}
\captionsetup{font=scriptsize}
\setlength{\tabcolsep}{1.5pt}
\begin{tabular}{@{}cc@{}}
\textbf{\scriptsize ChatDev 1.0} & \textbf{\scriptsize DevAll} \\[-0.2em]
\includegraphics[width=0.40\columnwidth]{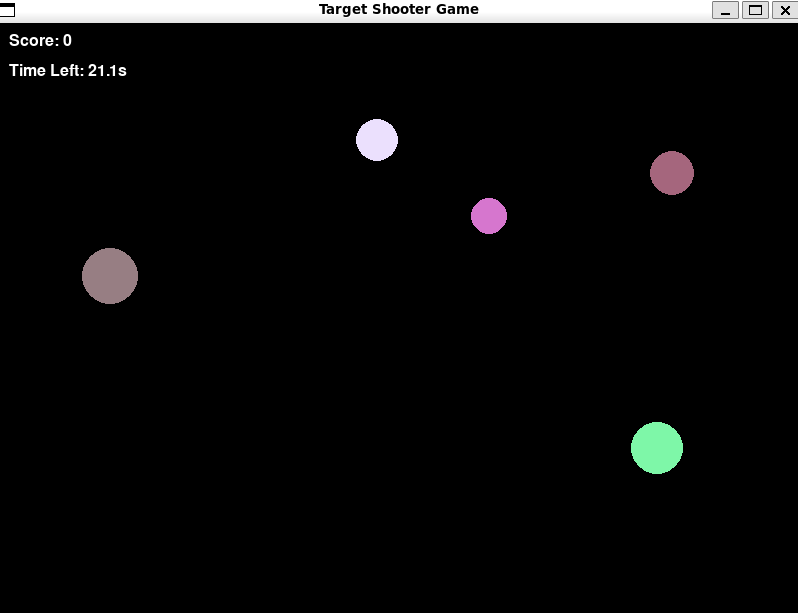}
&
\includegraphics[width=0.40\columnwidth]{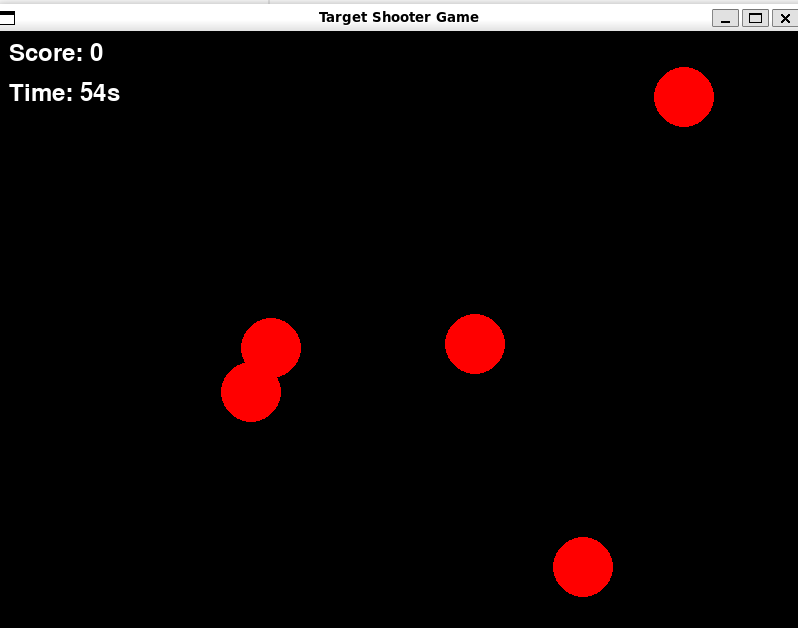}
\end{tabular}
\vspace{-0.8em}
\caption{Running screenshots of the target shooter game generated by ChatDev 1.0 and DevAll.}
\label{fig:app-software-case}
\vspace{-0.6em}
\end{figure}

We show representative outputs for visualization and software generation. As shown in Figures~\ref{fig:app-visualization-comparison} and~\ref{fig:app-software-case}, DevAll produces visible artifacts that are qualitatively comparable to the corresponding specialized baselines.

\FloatBarrier

\FloatBarrier

\end{document}